\documentclass{article}
\usepackage{spconf,amsmath,epsfig}
\usepackage{amsfonts}

\begin{document}\sloppy

% Example definitions.
% --------------------
\def\x{{\mathbf x}}
\def\L{{\cal L}}

\graphicspath{{./pictures/}}

% Title.
% ------
\title{BATCH FACE ALIGNMENT USING A LOW-RANK GAN}
%
% Single address.
% ---------------
\name{Jiabo Huang \qquad Xiaohua Xie\sthanks{Corresponding author. Email: xiexiaoh6@mail.sysu.edu.cn} \qquad Wei-Shi Zheng}
\address{School of Data and Computer Science, Sun Yat-sen University, China\\
         Guangdong Key Laboratory of Information Security Technology, China\\
         Key Laboratory of Machine Intelligence and Advanced Computing, Ministry of Education, China}

\maketitle

\begin{abstract}
This paper studies the problem of aligning a set of face images of the same individual into a normalized image while removing the outliers like partial occlusion, extreme facial expression as well as significant illumination variation. Our model seeks an optimal image domain transformation such that the matrix of misaligned images can be decomposed as the sum of a sparse matrix of noise and a rank-one matrix of aligned images. The image transformation is learned in an unsupervised manner, which means that ground-truth aligned images are unnecessary for our model. Specifically, we make use of the remarkable non-linear transforming ability of generative adversarial network(GAN) and guide it with low-rank generation as well as sparse noise constraint to achieve the face alignment. We verify the efficacy of the proposed model with extensive experiments on real-world face databases, demonstrating higher accuracy and efficiency than existing methods.
\end{abstract}

\section{Introduction}
\label{sec:intro}
Along with the rapid development of the internet in recent years, there are increasing number of image and video sharing sites such as Facebook and Youtube arise, which lead to a dramatic increase in the amount of human face data available online and also inspire lots of renewed interest in large, unconstrained face datasets \cite{huang2007labeled, wolf2011face}. However, domain transformations caused by significant illumination variation, partial occlusion, as well as poor or even no alignment make it difficult for most of the existing vision algorithms to work, such as the reconstruction of 3D face model and face recognition. Batch image alignment aims to align multiple images of an object or objects of interest to a fixed canonical template \cite{brown1992survey, maintz1998survey}. With the help of effective batch image alignment algorithm, unconstrained image set can be normalized and information encoded in them can be harnessed intelligently. In this work, we focus on batch face alignment and redefined this problem as aligning multiple face images from an individual to a fixed canonical template with the normalized poses, expressions, illumination conditions and occlusions.

To a large extent, progress in batch image alignment has been driven by the introduction of increasingly sophisticated measures of image similarity \cite{pluim2003mutual}. Some of the algorithms try to transform a misaligned matrix\footnote[1]{Flatten and stack a set of aligned/misaligned images as the columns of a matrix} to a rank one matrix in order to obtain a set of similar images. For examples, Learned-Miller's influential congealing algorithm \cite{learned2006data, huang2007unsupervised} aims to minimize the sum of entropies of pixel values at each pixel location while the least squares congealing procedure of \cite{cox2008least, cox2009least} seeks an alignment that minimizes the sum of squared distances between pairs of images. However, if there is a large illumination variation in images, the aligned matrix\footnotemark[1] might have an unknown rank higher than one. In this case, Vedaldi et. al. \cite{vedaldi2008joint} choose to minimize the rank of the aligned matrix. But such algorithms are unable to handle corruptions and occlusions that often occur in real images. Inspired by The Robust Parameterized Component Analysis (RPCA) algorithm of \cite{candes2011robust}, the Robust Alignment by Sparse and Low-rank Decomposition (RASL) \cite{peng2012rasl}, considering both large illumination variation and gross pixel corruptions, decomposes a misaligned matrix as the sum of a sparse noise matrix as well as a low-rank matrix of recovered aligned images. Overall, most of the existing batch image alignment algorithms solve the problem in a linear manner so it is difficult for them to handle non-linear variations such as illumination condition, partial occlusion as well as extreme facial expression.

GANs \cite{goodfellow2014generative}, as one of the most popular topics in recent few years, have been proved to have remarkable transforming ability between domains. The proposed adversarial loss, which guides the model to produce indistinguishable synthetic images, is the key to GANs' success. GANs have reached unprecedented heights in image generation \cite{denton2015deep,radford2015unsupervised} and image edition \cite{zhu2016generative}. Recently, they have also achieved impressive results in conditional image generation applications, such as text2image \cite{reed2016generative} and image inpainting \cite{pathak2016context}. In all these works, the distributions of both source and target domain are explicit, which means that there are numerous samples can be used by the GAN to fit the distributions of both sides. In this work, we prove that if a proper signal is given, GAN is capable of learning the transformation from an explicit source domain to an implicit target domain in an unsupervised way.

\textbf{Contributions.} In this paper, we propose an novel unsupervised learning model for robustly and efficiently aligning human face images, despiting partial occlusions and large variation of illumination. Our solution builds on recent advances in GAN, sparse decomposition, and low-rank analysis. It solves the batch face alignment problem by guiding the training of GAN with low-rank generation and sparse noise constraint. We show how GAN can be trained to fit an implicit distribution if a proper signal is given. We also verify the efficacy and efficiency of our model by experiments on real face images.

\textbf{Organization.} The remainder of this paper is organized as follows: In Section 2, we introduce how RASL works to align a set of images while how GAN works to transfer between domains and introduce our proposed method in details. Then we provide experimental results in Section 3 to showcase the efficacy of our model on real images. Section 4 provides concluding remarks and propose potential extensions to our model and the acknowledgements come last.

\section{Methodology}
\label{sec:method}
In this section, we present our method to batch face alignment. The proposed network is a variant of GAN supervised by low-rank generation and sparse noise constraint, which learns an effective transformation from the source domain to an implicit target domain.

\subsection{Sparse and Low-rank Decomposition}
It is general to measure the similarity of a set of images according to the rank of a matrix which is constructed by multiple images as columns. A well-aligned matrix should have low-rank because the images in it are linearly correlated.
However, In most practical scenarios, the low-rank matrix of correlated images breaks down easily if the images are even slightly misaligned with respect to each other or if there is any occlusions or corruptions in the images. 

RASL\cite{peng2012rasl} models the batch image alignment as a sparse and low rank decomposition which is formulated as Eq. \eqref{eq:RASL}
\begin{equation}
\label{eq:RASL}
\min \limits_{A,E,\tau} rank(A) + \gamma\lVert{E}\rVert_0\text{    s.t.    }D\circ\tau = A+E
\end{equation}
where $D$ represents the input misaligned matrix, $\tau : \mathbb{R}^2 \rightarrow \mathbb{R}^2$ is a set of invertible transformations such that $ I(x, y) = \{(I_j \circ \tau_j)(x,y)\}_{i=1}^n = \{I_j(\tau(x, y))\}_{i=1}^n $. Low rank matrix $A$ consists of $n$ well-aligned images while the sparse matrix $E$ represents the errors caused by occlusions and corruptions. $\gamma > 0$ is a parameter that trades off the rank of the solution versus the sparsity of the error. The objective searches for a set of transformations $\tau = {\tau_1, ..., \tau_n}$ such that the rank of the transformed images becomes as small as possible, when the sparse errors are subtracted. The low rank matrix provides us a well-aligned image set which is free of any occlusion and corruption.

In original RASL method, since the size of occlusions in different image sets can be totally different, it is necessary to adjust the $\gamma$ manually in order to balance the weights of low-rank constraint and sparse noise constraint according to different inputs. Besides, because the transformations are learned for each misaligned image separately, every time a new image set is fed into the model, RASL should searche for a set of transformations from scratch and this is quite inefficient. Therefore, both the robustness and efficiency of the RASL have a large room to be improved.

\subsection{GAN}
GANs enjoy a good reputation of being able to learn the between-domains transformation. It usually constructed by a generator and a discriminator. The generator aims to map the samples from the source domain into the target domain while the discriminator tries to distinguish between the real images from the target domain against the fake images produced by the generator. The objective of GAN can be formulated as:
\begin{equation}
\begin{split}
\min \limits_G \max \limits_D &\mathbb{E}_{x \in p_{data}(X)}[logD(x)]+\\
                              &\mathbb{E}_{z \in p_{data}(Z)}[log(1-D(G(z)))]
\end{split}
\end{equation}
where $x$ are images from target domain $X \in \mathbb{R}^{w \times h \times c}$ while $z$ are samples from source domain $Z \in \mathbb{R}^{w \times h \times c}$. $G$ tries to transfer the input $z$ into an image $G(z)$ which follows the distribution of target domain $X$ and $D$ produce the possibility of whether its input is a real sample from $X$.

GAN is able to learn the between-domains mapping only when there are a large amount of samples which can help the model determine the distributions of both domains. However, the target domain might be implicit in many problems. For example, for the batch image alignment, it is difficult to define the result of alignment for an input image set since even only one image in the input set is modified, the aligned result will be totally different. In such a problem, there are no ground-truth samples can help to determine target distribution. Besides, original GAN generally aims to learn an image-to-image mapping, but in the batch face alignment, we are trying to search for a set-to-set mapping instead. Therefore, although GAN shows its impressive ability of mapping between different domains, it is not intuitive to adopt it in our task.

\subsection{Low-rank GAN}
Inspired by RASL as well as GAN, our proposed model aims to take a set of misaligned images as input and generate a low-rank matrix as well as a sparse matrix of noise. We make the aligned matrix a rank-one matrix as \cite{cox2008least, cox2009least} did since the well-aligned images are supposed to be same in every aspect ideally. Therefore, we abandon the low-rank constraint and make the generator produce only one aligned image according to the whole misaligned image set. We keep the sparse noise constraint to make sure that the transformations made by GAN are actually doing the alignment as our expectation, which is related to all the input images. What's more, since the transformations of different images in an image set are supposed to be correlated, we concatenate fixed-length image set in the channels dimension as the input of the generator. The objective of our model is expressed as Eq. \eqref{eq:objective_sparse-gan}
\begin{equation}
\mathcal{L}_{Sparse}(G, T) = \mathbb{E}_{t \in p_{data}(T)}[\lVert t-G(t) \rVert_0]
\end{equation}
\begin{equation}
\begin{split}
\mathcal{L}_{GAN}(G, D, X, T)&=\mathbb{E}_{x \in p_{data}(X)}[logD(x)]\\
                    &+ \mathbb{E}_{t \in p_{data}(T)}[log(1-D(G(t)))]
\end{split}
\end{equation}
\begin{equation}
\label{eq:objective_sparse-gan}
\begin{split}
\mathcal{L}_{Full}(G, D, X, T)&= \mathcal{L}_{GAN}(G, D, X, T)\\
                       &+ \gamma\mathcal{L}_{sparse}(G, T)
\end{split}
\end{equation}
where $T \in \mathbb{R}^{w \times h \times c \times n}$ is a set of $n$ misaligned images from domain $X$ concatenated in the channels dimension. $G$ tries to synthesis image $G(t)$ that look similar to images from domain $X$ as well as the images in the input $t$, while $D$ aims to distinguish between synthetic image $G(t)$ and real samples $x$.  $\lVert t-G(t) \rVert_0$ is the $l^0\text{-norm}$ of the noise matrix and $\gamma$ controls the relative importance of adversarial objective and sparse noise constraint's objective. We aim to solve:
\begin{equation}
\label{eq:optimize_form}
\left\{G^*, D^*\right\}=\arg \min \limits_G \max \limits_D \mathcal{L}_{Full}(G, D, X, T)
\end{equation}
The optimization of Eq. \eqref{eq:optimize_form} is not directly tractable because of the nonconvexity of $l^0\text{-norm}$. It was shown that optimization of $l^0\text{-norm}$ can be replaced by optimizing $l^1\text{-norm}$ as long as the number of non-zero entries in matrix is not too large \cite{candes2011robust}. Therefore, the objective of sparse noise constraint can be modified into
\begin{equation}
\mathcal{L}_{Sparse}(G, T) = \mathbb{E}_{t \in p_{data}(T)}[\lVert t-G(t) \rVert_1]
\end{equation}
The gradient-based updates can use any standard gradient-based learning rule and we used standard Adam solver \cite{kingma2014adam} in our experiments.

\subsection{Network Architecture}
\begin{figure}[tb]
\begin{minipage}[b]{1.0\linewidth}
  \centering
  \centerline{\includegraphics[width=8.5cm]{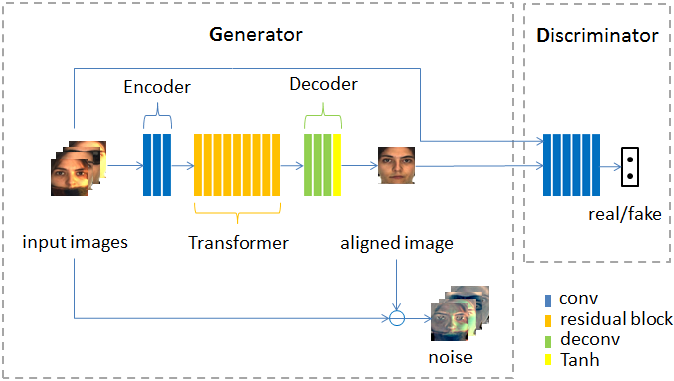}}
  % \centerline{(a) Result 1}\medskip
\end{minipage}
\caption{Network Structure. The network is constructed by a generator and a discriminator while the generator consists of an encoder, a transformer and a decoder. Best view in colored.}
\label{fig:structure}
\end{figure}

We adopt the network architecture from CycleGAN \cite{zhu2017unpaired} which has achieved impressive results in neural style transfer. The generator contains two stride-2 convolution layers, 9 residual blocks as well as two fractionally strided convolutions with stride $1/2$ for $160 \times 160$ input images. The discriminator is a $70 \times 70$ PatchGANs, since it aims to classify whether several overlapping patches from an image are real or fake in a fully convolutional manner, it can be applied to arbitrarily-sized inputs. And such a patch-level discriminator has fewer parameters than a full-image discriminator. Instance normalization is also used during training. The overall network structure is shown in 
Fig. \ref{fig:structure}. The generator is constructed by an encoder, a transformer as well as a decoder. It takes multiple images from an individual as input and concatenates them in channels dimension before feeding them to the encoder and finally produces a synthetic image with the same size as input. After that, both input images and synthetic images are fed to the discriminator and the discriminator will then produce the possibilities of whether they are real or fake images.

\section{Experiments}
\label{sec:exp}
We evaluate our model on AR Face Database \cite{martinez2007ar}. This face database contains over 4,000 color images from 70 men and 56 women. All the images contain one frontal face with different facial expressions, illumination conditions and occlusions (sun glasses and scarf). We separate the database into a training set and a testing set containing 116 and 10 people respectively. We first compare the aligned results of RASL with our model to prove that our model is more robust to a large area of occlusion and extreme variation of expression. Then, we try to reconstruct the 3D face model according to the aligned results of both models. Through this experiment, we aim to show that a high-quality aligned image is significant for face image processing like 3D face reconstruction.

\subsection{Implementation Details}
For all input images, we first crop the face region with the help of the face detector MTCNN \cite{zhang2016joint}, then we resize them into $160 \times 160$. Each image is standardized with zero means and unit standard deviation. We randomly separate all images from each individual into several image sets and each set contains 8 images. We replace the negative log likelihood objective by a least-squares loss \cite{mao2016multi}. This loss is more stable during training and can help the model generate higher quality results. We set $\gamma=2e^{-5}$ and use the standard Adam solver with a batch size of 16. All networks are trained from scratch with a fixed learning rate of 0.0002. Our model is implemented by tensorflow and we do all the experiments using a standard PC with a NVIDIA Titan Pascal GPU.

\subsection{Aligned Results}
\begin{figure}[tb]
\begin{minipage}[b]{1.0\linewidth}
  \centering
  \centerline{\includegraphics[width=8.5cm]{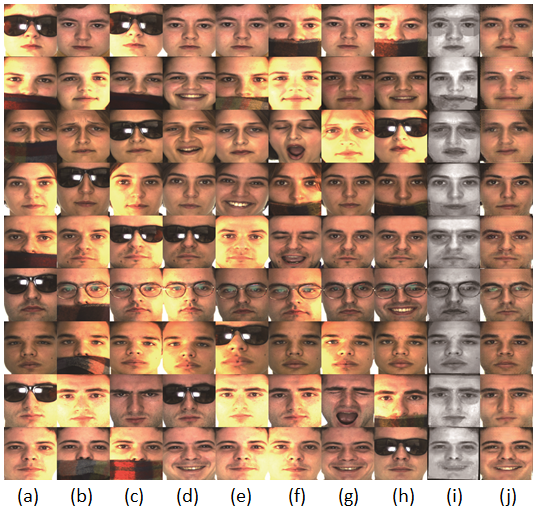}}
  % \centerline{(a) Result 1}\medskip
\end{minipage}
\caption{Comparison of the aligned results from RASL and our model. Each row shows samples from one individual in the testing set. Columns (a)-(h) are the input images, (i) is the aligned result from RASL while (j) is the result of our model.}
\label{fig:results}
\end{figure}

For RASL, we leave all the parameters set in default and take the average of images, which are contained in the produced low-rank matrix, as its output. Both the input and output of low-rank decomposition in RASL is in grayscale. Fig. \ref{fig:results} shows a few examples from the testing set. Comparing the outputs in the second row, when there is a large area of occlusion like a scarf, RASL cannot perfectly remove them but the situation is better in our model. Comparing the outputs in the third row and the last row, when there is a large variation of expression in the input images, the mouth region in RASL's output is blurred while ours is sharper.

\subsection{3D Reconstruction}
\begin{figure}[tb]
\begin{minipage}[b]{1.0\linewidth}
  \centering
  \centerline{\includegraphics[width=8.5cm]{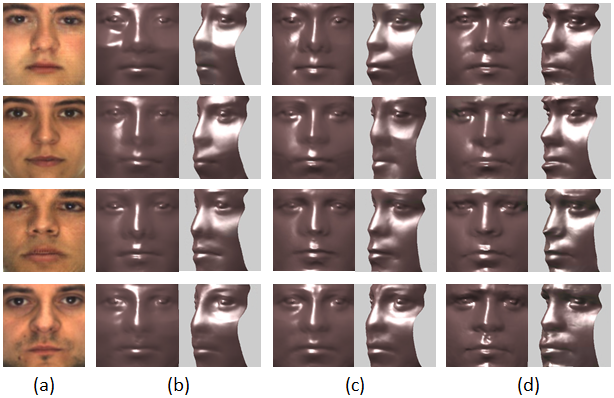}}
  % \centerline{(a) Result 1}\medskip
\end{minipage}
\caption{Comparison of the 3D reconstructed results. (a) is sample images from input. The complete input image sets are shown in Fig. \ref{fig:results}. (b), (c) and (d) compare the 3D reconstructed results of original images, aligned images from RASL and our model respectively.}
\label{fig:3d-reconstruction}
\end{figure}

Hu et. al. proposed a 3D reconstruction algorithm \cite{hu2017sparse} which takes $n \geq 1$ frontal face images from an individual to reconstruct the 3D face model. We make use of it and carefully set all the parameters as well as the directions of illumination in each image as it requires. For each individual, we use 8 original images, 8 RASL's aligned images as well as one aligned image from our model for the reconstruction. From Fig. \ref{fig:3d-reconstruction} (b), it is clear that occlusion will certainly affect the quality of the reconstructed 3D model. The second row of Fig. \ref{fig:3d-reconstruction} (c) shows that aligned face from RASL with partial blur can sometimes make the situation worse while (d) proves that our aligned result, encoding all the useful information and free of any occlusion, does help the reconstruction.

\section{Conclusion}
\label{sec:conclusion}
We propose an effective batch face alignment method based on GAN and use the low rank and sparse constraint to supervise the training of our model in order to prove that GAN is capable of learning a mapping to an implicit domain. Our model is more efficient than traditional matrix decomposition based method since after the off-line training, it doesn't need to learn transformations from scratch when a new image set is fed. And the efficacy of our model is also verified by the experiments. In the future, we would like to consider the case of poses variation.

\subsubsection*{Acknowledgments}
\label{sec:ack}
This project is supported by the NSFC (No. U1611461, 61672544), Guangdong Natural Science Foundation (No. 2015A030311047), Fundamental Research Funds for the Central Universities (No. 161gpy41), and Tip-top Scientific and Technical Innovative Youth Talents of Guangdong special support program (No. 2016TQ03X263).

\bibliographystyle{IEEEbib}
\bibliography{refs}

\end{document}